\documentclass[11pt]{article}
\usepackage{coling2018}
\usepackage{times}
\usepackage{pstricks}
\usepackage{url}
\usepackage{latexsym}

\usepackage{makecell}
\usepackage[T1]{fontenc}
\usepackage{CJKutf8}
\usepackage[english]{babel}
\usepackage{amsmath,graphicx}
\usepackage{listings}
\usepackage{tabularx}

\usepackage{pgfplots}
\pgfplotsset{width=7cm,compat=1.9}
\slname{Korean}

\title{Rich Character-Level Information for Korean Morphological Analysis and Part-of-Speech Tagging}

\author{Andrew Matteson, Chanhee Lee, Heuiseok Lim\thanks{\hspace{0.1cm} Corresponding author} \\
  Korea University \\
  {\tt \{amatteson, chanhee0222, limhseok\}@korea.ac.kr} \\ \\
  \textbf{Young-Bum Kim} \\
  Amazon Alexa\\
  {\tt youngbum@amazon.com}}

\date{}

\begin{document}
\maketitle
\begin{abstract}
  Due to the fact that Korean is a highly agglutinative, character-rich language, previous work on Korean morphological analysis typically employs the use of sub-character features known as graphemes or otherwise utilizes comprehensive prior linguistic knowledge (i.e., a dictionary of known morphological transformation forms, or actions). These models have been created with the assumption that character-level, dictionary-less morphological analysis was intractable due to the number of actions required. We present, in this study, a multi-stage action-based model that can perform morphological transformation and part-of-speech tagging using arbitrary units of input and apply it to the case of character-level Korean morphological analysis. Among models that do not employ prior linguistic knowledge, we achieve state-of-the-art word and sentence-level tagging accuracy with the Sejong Korean corpus using our proposed data-driven Bi-LSTM model.
\end{abstract}


\section{Introduction}
\label{intro}

%
%
\blfootnote{
    %
    %
    %
    %
    %
    %
    \hspace{-0.65cm}  
    This work is licensed under a Creative Commons 
    Attribution 4.0 International License.
    License details:
    \url{http://creativecommons.org/licenses/by/4.0/}
}

Korean has traditionally posed a challenge for word segmentation and morphological analysis. In addition to virtually unbounded vocabulary sizes, out-of-vocabulary (OOV) rates for models can be high. Korean is an agglutinative, phonetic language with a SOV (Subject-Object-Verb) syntax and a flexible word order, although certain word orders are considered to be ``canonical''. Honorifics, conditionals, imperatives, and other forms are all signified using agglutinative endings which sometimes involve transformation of the stem to which they attach. Some endings can be further combined or fused to other endings in a defined order, and furthermore, morphological transformation rules also apply during this process. Transformation rules are mostly consistent at the grapheme level and can be represented by a handful of spelling rules, but many irregular forms do exist.

In Unicode, Hangul (Korean alphabet) characters are allocated 11,140 codepoints. Each character contains an initial consonant, vowel, and final consonant represented in C+V+C, C+V, or V+C form. In Unicode, the form is always assumed to be C+V+C and the initial or final consonants are set to null according to the desired target form. Consonants and vowels are considered to be sub-character units called graphemes. Each character is represented using a combination of 19 initial consonants (including null), 21 vowels, and 27 final consonants (including null), and there is a mathematical formula that can be used to combine graphemes to generate the codepoint of a Hangul character. The character ``{\begin{CJK}{UTF8}{mj}김\end{CJK}}'' (gim) can be represented in C+V+C form as follows.
\bigskip
\begin{center}
{%
\small
{\centering
\begin{tabular}{|c|c|c|}
\hline \bf \makecell{Initial \\ Consonant} & \bf Vowel & \bf \makecell{Final \\ Consonant} \\ \hline
{\begin{CJK}{UTF8}{mj}ㄱ\end{CJK}}(g) & {\begin{CJK}{UTF8}{mj}ㅣ\end{CJK}}(i) & {\begin{CJK}{UTF8}{mj}ㅁ\end{CJK}}(m) \\
\hline
\end{tabular}
}}
\end{center}

\bigskip
The Korean language has ``fusion'' spelling rules that apply across character boundaries (within an agglutinative unit), which implies that morphological transformation may occur among adjacent graphemes. When the final consonant of one character meets the initial consonant of the next character during verb inflection, there may be a change in the resulting combined character. This presents character-level embeddings with a unique challenge that is not present in most other languages.

\begin{table}[!t]
\parbox{.4\linewidth}{
\centering
{%
\begin{tabular}{|l|c|}
\hline \bf \# & \bf Morphemes \\ \hline
1 & {\begin{CJK}{UTF8}{mj}나\end{CJK}}(na)/VV + {\begin{CJK}{UTF8}{mj}는\end{CJK}}(neun)/ETM \\ \hline
2 & {\begin{CJK}{UTF8}{mj}날\end{CJK}}(nal)/VV + {\begin{CJK}{UTF8}{mj}는\end{CJK}}(neun)/ETM \\ \hline
3 & {\begin{CJK}{UTF8}{mj}나\end{CJK}}(na)/NP + {\begin{CJK}{UTF8}{mj}는\end{CJK}}(neun)/JX \\ \hline
4 & {\begin{CJK}{UTF8}{mj}나\end{CJK}}(na)/NNP + {\begin{CJK}{UTF8}{mj}는\end{CJK}}(neun)/JX \\ \hline
5 & {\begin{CJK}{UTF8}{mj}나\end{CJK}}(na)/VX + {\begin{CJK}{UTF8}{mj}는\end{CJK}}(neun)/ETM \\ \hline
\end{tabular}
}
\caption{\label{ambiguous-parse} Ambiguous parses of Eojeol ``na-neun''}
}
\hfill
\parbox{.55\linewidth}{
\centering
{%
\begin{tabular}{|c|c|c|}
\hline \bf \# & \bf Eojeol & \bf Morphemes \\ \hline
(3) & \makecell{{\begin{CJK}{UTF8}{mj}나는\end{CJK}} \\ na-neun} & {\begin{CJK}{UTF8}{mj}나\end{CJK}}(na)/NP + {\begin{CJK}{UTF8}{mj}는\end{CJK}}(neun)/JX \\ \hline
 & \makecell{{\begin{CJK}{UTF8}{mj}하늘에\end{CJK}} \\ ha-neul-e} & {\begin{CJK}{UTF8}{mj}하늘\end{CJK}}(ha-neul)/NNG + {\begin{CJK}{UTF8}{mj}에\end{CJK}}(e)/JKB \\ \hline
(2) & \makecell{{\begin{CJK}{UTF8}{mj}나는\end{CJK}} \\ na-neun} & {\begin{CJK}{UTF8}{mj}날\end{CJK}}(nal)/VV + {\begin{CJK}{UTF8}{mj}는\end{CJK}}(neun)/ETM \\ \hline
 & \makecell{{\begin{CJK}{UTF8}{mj}새를\end{CJK}} \\ sae-leul} & {\begin{CJK}{UTF8}{mj}새\end{CJK}}(sae)/NNG + {\begin{CJK}{UTF8}{mj}를\end{CJK}}(leul)/JKO \\ \hline
 & \makecell{{\begin{CJK}{UTF8}{mj}보았다\end{CJK}} \\ bo-ass-da.} & \makecell{{\begin{CJK}{UTF8}{mj}보\end{CJK}}(bo)/VV + {\begin{CJK}{UTF8}{mj}았\end{CJK}}(ass)/EP \\ + {\begin{CJK}{UTF8}{mj}다\end{CJK}}(da)/EF + ./SF} \\
\hline
\end{tabular}
}
\caption{\label{ambiguous-sentence} Correct transformation and tag sequence for sample sentence containing ambiguous Eojeol ``na-neun''. \# corresponds to correct parse sequence in Table~\ref{ambiguous-parse}, only labeled here for ``na-neun''.}
}
\end{table}

In order to avoid confusion of terminology, we must define the precise meaning of morphological analysis in the context of Korean. For most languages, morphological analysis refers to a word-level tag that describes the aspect, tense, plurality, and other features of the word, whereas part-of-speech (POS) tagging serves to classify the word as a noun, verb, etc. The POS tag is sometimes concatenated to the morphological tag string as in the POSMORPH annotation employed by Heigold, et al~\shortcite{HeigoldOriginalModel}.

In Korean, morphological analysis refers to the segmentation and restoration of morphemes within a ``word'' unit called an Eojeol and the POS tagging of each constituent morpheme. An Eojeol encodes not only lexical information but also grammatical information due to the agglutinative nature of the Korean language. The recovered morpheme segments often include a stem and other morphemes which indicate tense or other linguistic features. Traditional Korean morphological analysis algorithms operate at the Eojeol level and yield all ambiguous parses (Table~\ref{ambiguous-parse}) that lead to that particular Eojeol, including the morpheme transformations and tags. However, the model\footnote{Model source code is made available at https://github.com/xtknight/rich-morphological-tagger} proposed in this paper receives input at the sentence level and attempts to produce the one correct sequence of transformations and tags for all Eojeol within the sentence according to the context (Table~\ref{ambiguous-sentence}).

\section{Related Work}

Morphological analysis of the Korean language has traditionally been performed in several ways, including separation of Korean characters into graphemes by using linguistic knowledge, lattice tree lookup~\cite{Park:2010}, application of regular and irregular inflection rules~\cite{Kang:1992}, morphosyntactic rule sets, and by using a pre-computed dictionary~\cite{Shim:2004}. However, we investigate whether morphological analysis of Korean is feasible without the use of any of these techniques and without a dictionary by making the assumption that common transformations and their underlying grapheme modifications can be easily recognized and learned with a Bi-LSTM model.

Bi-LSTM-CRFs have been used for sequential tagging with BIO annotation ~\cite{Sang:1999}. Huang, et al~\shortcite{Huang:2015} show their effectiveness for POS tagging, chunking, named entity recognition (NER). These models show state-of-the-art accuracy at several tasks.

Similar models have also been proposed in universal morphological analysis. Heigold, et al~\shortcite{HeigoldUniversalModel} show how a nested LSTM architecture can be applied to word-level morphological tagging for a wide variety of languages. At the lower level, an LSTM network is used for character-level embedding to reduce OOV errors. However, this work does not investigate how such a model would operate for the most widely used Korean Sejong Corpus.

Sub-character tagging has also been attempted. Dong, et al~\shortcite{Dong:2016} demonstrate how radical-level features incorporated at the character-level for named entity recognition achieve state-of-the-art accuracy for Chinese. The most convincing attempt to tag Korean at the morpheme level is by Choi, et al~\shortcite{Choi:2016} who achieve state-of-the-art (dictionary-less) performance by using a multi-stage Bi-LSTM-CRF model that involves the splitting of Korean character input into constituent graphemes. However, the implicit assumption that Korean characters must first be split into graphemes to achieve optimal performance for morphological analysis is not well supported, and we should consider the splitting of characters into graphemes to be employing linguistic knowledge specific to Korean.

In our paper, we seek to answer the question of whether Korean morphemes can be tagged without grapheme-level splitting, rules specific to the language, or a dictionary. Although we initially considered Bi-LSTM-CRF for our model architecture, we show that the performance benefit by adding CRF is minimal and practically unnecessary compared to a standard Bi-LSTM model. Furthermore, CRF adds training and inference computational complexity due to the Viterbi algorithm.

\section{Lemma and Form Alignment}

\begin{figure}[!htb]
\parbox{.55\linewidth}{
\small
\centering
{%
\begin{tabular}{|c|c|c|c|c|}
\hline {\begin{CJK}{UTF8}{mj}고\end{CJK}}(go) & {\begin{CJK}{UTF8}{mj}통\end{CJK}}(tong) & {\begin{CJK}{UTF8}{mj}스\end{CJK}}(seu) & \multicolumn{2}{c|}{{\begin{CJK}{UTF8}{mj}런\end{CJK}}(reon)} \\ \hline
B-KEEP & I-KEEP & B-KEEP & \multicolumn{2}{c|}{I-MOD-{\begin{CJK}{UTF8}{mj}럽\end{CJK}}, B-MOD-{\begin{CJK}{UTF8}{mj}ㄴ\end{CJK}}} \\ \hline
\multicolumn{2}{|c|}{{\begin{CJK}{UTF8}{mj}고통\end{CJK}}(go-tong)} & \multicolumn{2}{c|}{{\begin{CJK}{UTF8}{mj}스럽\end{CJK}}(seu-reob)} & {\begin{CJK}{UTF8}{mj}ㄴ\end{CJK}}(n) \\ \hline

\end{tabular}
}
\caption{\label{sejong-gold-morph-example-table} Gold morphological transformation actions given by alignment oracle, including the resulting morphemes after running the BIO actions (Sejong corpus)}
}
\hfill
\parbox{.4\linewidth}{
\small
\centering
{%
\begin{tabular}{|c|c|c|c|c|}
\hline \multicolumn{2}{|c|}{{\begin{CJK}{UTF8}{mj}고통\end{CJK}}(go-tong)} & \multicolumn{2}{c|}{{\begin{CJK}{UTF8}{mj}스럽\end{CJK}}(seu-reob)} & {\begin{CJK}{UTF8}{mj}ㄴ\end{CJK}}(n) \\ \hline
\multicolumn{2}{|c|}{NNG} & \multicolumn{2}{c|}{XSA} & ETM \\ \hline
\end{tabular}
}
\caption{\label{sejong-gold-tag-example-table} Gold tagging actions (Sejong corpus)}
}
\end{figure}

In the Sejong corpus, Eojeol are annotated with their corresponding POS-tagged morpheme constituents, exactly as shown in Table~\ref{ambiguous-sentence}. As mentioned earlier, morpheme spelling transformations may occur, and therefore the morpheme constituents may have slightly different graphemes than what is present in the original Eojeol form. To generate our training data, we must align the Eojeol form and its constituent morphemes at the character level, as we forbid using linguistic knowledge such as sub-character elements (graphemes) in our model. Like most agglutinative languages, the Eojeol form and lemmas (morphemic elements in the Sejong corpus) often share overlapping characters at the beginning or end, and we utilize this assumption in our algorithm.

\begin{table}[!htb]
\bigskip
\centering
{%
\begin{tabular}{|c|c|}
\hline \bf Gold Action & \bf Count \\
\hline B-KEEP & 23,725,534 \\
\hline I-KEEP & 8,650,166 \\
\hline B-MOD-{\begin{CJK}{UTF8}{mj}하\end{CJK}}(ha), B-MOD-{\begin{CJK}{UTF8}{mj}ㄴ\end{CJK}}(n) & 153,130 \\
\hline NOOP & 131,016 \\
\hline B-MOD:{\begin{CJK}{UTF8}{mj}하\end{CJK}}(ha), B-MOD-{\begin{CJK}{UTF8}{mj}았\end{CJK}}(ass) & 61,592 \\
\hline B-MOD:{\begin{CJK}{UTF8}{mj}이\end{CJK}}(i), B-MOD-{\begin{CJK}{UTF8}{mj}ㄴ\end{CJK}}(n) & 58,093 \\
\hline B-MOD:{\begin{CJK}{UTF8}{mj}하\end{CJK}}(ha), B-MOD-{\begin{CJK}{UTF8}{mj}아\end{CJK}}(a) & 57,515 \\
\hline I-MOD:{\begin{CJK}{UTF8}{mj}하\end{CJK}}(ha), B-MOD-{\begin{CJK}{UTF8}{mj}아\end{CJK}}(a) & 48,987 \\
\hline B-MOD:{\begin{CJK}{UTF8}{mj}되\end{CJK}}(doe), B-MOD-{\begin{CJK}{UTF8}{mj}ㄴ\end{CJK}}(n) & 41,335 \\
\hline B-MOD:{\begin{CJK}{UTF8}{mj}하\end{CJK}}(ha), B-MOD-{\begin{CJK}{UTF8}{mj}ㄹ\end{CJK}}(l) & 36,419 \\ \hline
\end{tabular}
}
\caption{\label{common-actions} Top 10 morphing actions}
\end{table}

We present an action-based algorithm (called an ``alignment oracle'') to align two arbitrary strings. Our oracle attempts to generate a 1:1 character-level mapping between the morphological form and the lemmas of an Eojeol by searching for a prefix, suffix, and modified inner string portion. Three primary actions are defined: KEEP (no modification to character), NOOP (drop character), and MOD (modify character). In the case of Korean, morphological transformations happen at the end of a form, so there is rarely a common suffix unless no transformation occurs at all. These primary actions are then augmented with B- and I- actions to facilitate morpheme segmentation. It is important to note that our algorithm is not specific in any way to the Sejong corpus or Korean itself.

The full process is demonstrated in Figure~\ref{sejong-gold-morph-example-table} starting from the source form. The gold untagged segmented lemma form is shown in the bottom row, and the actions generated by the oracle to generate the lemma are given in the middle row. The first three characters (go-tong-seu) are preserved with KEEP actions and the last character is considered the ``modified inner string''. In this case, the number of actions (5) exceeds the number of full input characters (4), and therefore two actions are assigned to the last character which split the ``reon'' syllable into ``reob'' and ``n''. The output after morpheme segmentation can be seen in the bottom row. In Figure~\ref{sejong-gold-tag-example-table}, these output morphemes are then placed through a standard sequential tagger to assign part-of-speech tags.

\begin{figure}[!htb]
    \centering
    \def\svgwidth{\columnwidth}    
    \scalebox{1.0}{\input{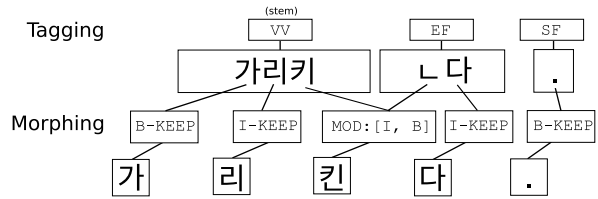}}
	\caption{\label{stage-actions} Actions for transformation with output segmentation }
\end{figure}

For Korean, the task is considerably more complicated. Rather than merely character-level transformation, new morpheme boundaries based on the results of those transformations are also required. A segmentation module adds B-/I- (beginning and inside) annotations to the KEEP and MOD actions. These actions allow morpheme segmentation to take place even amidst the modified character output sequence. This is detailed in Figure~\ref{sejong-gold-morph-example-table}, where the final consonant sub-character unit (``n'') of the last character of input (``reon'') is transformed to ``b'' and the resulting fused full character is appended to the previous output morpheme, whereas the ``n'' sub-character unit becomes separated and represented as an entirely new morpheme itself. The top 10 resulting actions for the Sejong corpus on the form and lemma alignment stage are shown in Table~\ref{common-actions}.

\begin{figure*}
    \centering
    \def\svgwidth{\columnwidth}    
    \scalebox{1.5}{\input{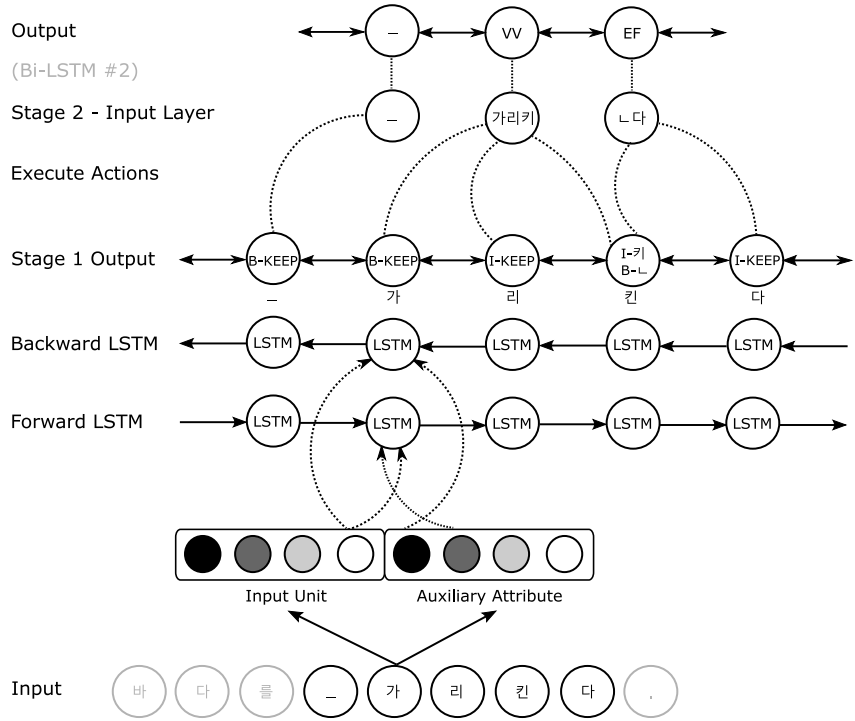}}
	\caption{\label{bi-lstm-fig} Partial example of two-stage tagging process for Korean phrase ``to point out the sea''}
\end{figure*}

\section{The Model}

Our model makes the assumption that in order to support morphological analysis for languages like Korean, two stages are required, which we call morphing and tagging. For tasks such as morphological transformation, word-level morphological analysis, or morpheme segmentation, only one stage is strictly necessary. To obtain tags for morphemes following morphological transformation, as in Korean, both stages are necessary, with tagging following morphing. The stages have no fundamental difference from each other: the second stage simply acts on the result of applying actions output by the first stage. Each stage outputs a single action for a single input unit. A single action could be as simple as a tag or as complicated as information resulting in advanced multi-character transformation along with the specification of the morphemic segments of those resulting characters. Model parameters are trained independently for each stage unless otherwise specified.

The model presented in this paper is inspired by word-level morphological analysis work by Heigold, et al~\shortcite{HeigoldOriginalModel} with the goal of allowing analysis at arbitrary units of input at each stage. Because we do not specify whether the input unit should be a word, morpheme, character, or even a unit at the sub-character level (such as graphemes in Korean or radicals in Chinese), we theoretically have the flexibility to tag a variety of languages at any level.

We employ a bidirectional Long Short-Term Memory \cite{Hochreiter:1997,Graves:2005} network in our model and also experiment with optimization to a conditional random field (CRF) objective~\cite{Laffety:2001}. In theory, CRF allows us to consider the likelihood of neighboring outputs and therefore jointly decode the highest probable chain of output labels for a given set of inputs. Although we posit that CRF is not strictly necessary, the overall architecture of our model is otherwise identical to a standard Bi-LSTM-CRF sequence tagging model \cite{Huang:2015} used for POS tagging and NER.

An overview of the architecture is shown in Figure~\ref{bi-lstm-fig}. The input unit is embedded as a multi-dimensional vector. At the input level, an auxiliary attribute may be concatenated with the input unit to include auxiliary information for the unit, such as word break-level information, although empirical findings indicate our model performs best when using only the input unit. Nested embeddings as in~\shortcite{HeigoldOriginalModel} may also be appended at the input level.

All embeddings are concatenated to form a combined embedding which is then passed to the primary Bi-LSTM-CRF network and trained against a set of output actions. Whitespace delimiting Eojeols is represented as a reserved spacing token in the input unit.

Although each stage is independent and can accept an arbitrary unit of input suitable for any language, the following sections describe how this model pertains to our primary task of morpheme-level morphological analysis for the Korean language.

\subsection{Morphing}

The first stage of the model operates at the character level and is responsible for morphological transformation of the input form into the desired output lemma(s). During training, each input character is assigned one of three primary types by the alignment oracle as described in Section 3. Morpheme segmentation actions are also generated and augmented to the transformation actions at this stage for proper morpheme boundary identification. During inference, instead of using the alignment oracle, one action (including transformation and B-/I- tags) is predicted for each character based on trained parameters. At this point, tags are not yet assigned to each output morpheme.

\subsection{Tagging}

After the necessary morphological transformation and segmentation, tagging occurs at the morpheme level and acts on output produced by actions in the first stage of the model. An example of this is shown in Figure~\ref{sejong-gold-tag-example-table}. In this stage, the action is simply to assign a POS tag to the morpheme, which is the input unit.

\section{Experiments}

\subsection{Datasets}

We conduct experiments using the full Sejong Korean Balanced Corpus dataset. The experiments are coded in Python using the TensorFlow library. The Sejong Corpus has been preprocessed to resolve punctuation inconsistencies and other surface-level errors. All datasets are converted at the sentence-level to a simple two-column format with each line containing an input unit and target action.

For all experiments, we follow an 85/10/5 cross-validation split for training, testing, and validation sets respectively. All data is randomly shuffled prior to splitting. Actions are inferred from the dataset by using lemma and form alignment. For evaluation, output from predicted actions in the first stage is used as input to the subsequent stage.

$UniTagger$ represents the model proposed in this paper. The following number (for example, 500) represents the maximum action count for the morphing stage. The tagging stage only has as many actions as possible POS tags (45 in the Sejong corpus, including the reserved space token). Action pruning is performed at the training level, which removes from the training set the least common morphological transformation actions generated by the alignment oracle. For fair evaluation, actions are not removed from validation or test sets. All accuracy figures in this paper are reported based on a held-out test set.

\subsection{Training}

Optimization is performed using Adam~\cite{Kingma:2014} with a learning rate of 0.001 and decay of 0.9. In the case of multi-stage models, model parameters are optimized independently for each stage. We use identical hyperparameters for all morphing and tagging models. Input unit embedding size was set to 300 (for character and morpheme input). The final Bi-LSTM concatenating all embeddings before an optional CRF layer was used with an LSTM unit size of 300. Batch size was set to 64 for all experiments, except for the CRF experiment where it was set to 16. The maximum LSTM input length was set at a per-batch level which yielded optimal performance, and the maximum number of input units (whether characters or morphemes) was limited to 400 in both stages. A dropout of 10\% was used for the reported model with best performance. Dropout is only applied at the unit embedding layer. Epoch count was set to 100 with early-stopping after 3 epochs with no improvement in validation set performance. Experiments were performed on GTX 1080 Ti 11GB GPUs. Average total training duration was around 5 hours for the entire Sejong dataset on a GTX 1080 Ti. In TensorFlow, the NVIDIA CuDNN-optimized LSTM was used~\cite{Appleyard:2016}.

\subsection{Results}

In Table~\ref{wordlevel-table}, we show Eojeol-level morphological analysis accuracy for Korean. Note here that an Eojeol is considered correctly tagged only if all its constituent morphemes have been transformed, segmented, and tagged properly. Table~\ref{sentencelevel-table} measures sentence-level tagging performance, which is the accuracy of all morphemes being transformed and tagged properly.

\begin{table}[!htb]
\parbox{.45\linewidth}{
\centering
\begin{tabular}{|l|c|}
\hline \bf Model & \bf Accuracy \\ \hline
Lee, et al.~\shortcite{Lee1} & 92.96 \\
Ahn, et al.~\shortcite{Ahn:2007} & 93.12 \\
Lee, et al.~\shortcite{Lee2} & 92.95 \\
Choi, et al.~\shortcite{Choi:2016} & 94.89 \\
UniTagger-500 & \textbf{96.20} \\
\hline
\end{tabular}
\caption{\label{wordlevel-table} End-to-end Eojeol-level accuracy for morphological analysis of Korean (Sejong Corpus)}

\bigskip
\bigskip
\centering
\begin{tabular}{|l|c|c|}
\hline \bf Model & \bf Model Type & \bf Acc \\ \hline
Choi, et al.~\shortcite{Choi:2016} & Bi-LSTM-CRF & 61.00 \\
UniTagger-500 & Bi-LSTM & \textbf{70.83} \\
\hline
\end{tabular}
\caption{\label{sentencelevel-table} End-to-end sentence-level accuracy for morphological analysis of Korean (Sejong Corpus)}

}
\hfill
\parbox{.45\linewidth}{
\centering
{%
\begin{tabular}{|c|c|}
\hline \bf Form & \bf Gold Action \\
\hline {\begin{CJK}{UTF8}{mj}났\end{CJK}}(nass) & B-{\begin{CJK}{UTF8}{mj}나\end{CJK}} + B-{\begin{CJK}{UTF8}{mj}았\end{CJK}} \\
\hline {\begin{CJK}{UTF8}{mj}샀\end{CJK}}(sass) & B-{\begin{CJK}{UTF8}{mj}사\end{CJK}} + B-{\begin{CJK}{UTF8}{mj}았\end{CJK}} \\
\hline {\begin{CJK}{UTF8}{mj}잤\end{CJK}}(jass) & B-{\begin{CJK}{UTF8}{mj}자\end{CJK}} + B-{\begin{CJK}{UTF8}{mj}았\end{CJK}} \\
\hline {\begin{CJK}{UTF8}{mj}팠\end{CJK}}(pass) & B-{\begin{CJK}{UTF8}{mj}파\end{CJK}} + B-{\begin{CJK}{UTF8}{mj}았\end{CJK}} \\
\hline {\begin{CJK}{UTF8}{mj}됐\end{CJK}}(daess) & B-{\begin{CJK}{UTF8}{mj}되\end{CJK}} + B-{\begin{CJK}{UTF8}{mj}었\end{CJK}} \\
\hline {\begin{CJK}{UTF8}{mj}했\end{CJK}}(haess) & B-{\begin{CJK}{UTF8}{mj}하\end{CJK}} + B-{\begin{CJK}{UTF8}{mj}았\end{CJK}} \\
\hline {\begin{CJK}{UTF8}{mj}녔\end{CJK}}(nyeoss) & B-{\begin{CJK}{UTF8}{mj}니\end{CJK}} + B-{\begin{CJK}{UTF8}{mj}었\end{CJK}} \\
\hline {\begin{CJK}{UTF8}{mj}렸\end{CJK}}(ryeoss) & B-{\begin{CJK}{UTF8}{mj}리\end{CJK}} + B-{\begin{CJK}{UTF8}{mj}었\end{CJK}} \\
\hline {\begin{CJK}{UTF8}{mj}셨\end{CJK}}(syeoss) & B-{\begin{CJK}{UTF8}{mj}시\end{CJK}} + B-{\begin{CJK}{UTF8}{mj}었\end{CJK}} \\
\hline {\begin{CJK}{UTF8}{mj}졌\end{CJK}}(jieoss) & B-{\begin{CJK}{UTF8}{mj}지\end{CJK}} + B-{\begin{CJK}{UTF8}{mj}었\end{CJK}} \\
\hline {\begin{CJK}{UTF8}{mj}겼\end{CJK}}(gyeoss) & B-{\begin{CJK}{UTF8}{mj}기\end{CJK}} + B-{\begin{CJK}{UTF8}{mj}었\end{CJK}} \\
\hline {\begin{CJK}{UTF8}{mj}왔\end{CJK}}(oass) & B-{\begin{CJK}{UTF8}{mj}오\end{CJK}} + B-{\begin{CJK}{UTF8}{mj}았\end{CJK}} \\
\hline {\begin{CJK}{UTF8}{mj}놨\end{CJK}}(noass) & B-{\begin{CJK}{UTF8}{mj}놓\end{CJK}} + B-{\begin{CJK}{UTF8}{mj}았\end{CJK}} \\
\hline {\begin{CJK}{UTF8}{mj}췄\end{CJK}}(chuweoss) & B-{\begin{CJK}{UTF8}{mj}추\end{CJK}} + B-{\begin{CJK}{UTF8}{mj}었\end{CJK}} \\ \hline
\end{tabular}
}
\caption{\label{past-tense-actions} Past tense morphing actions shown in embedding and gold actions from Sejong corpus}
}
\end{table}

\begin{table}[!htb]
\bigskip
\centering
\begin{tabular}{|l|c|c|}
\hline \bf Model & \bf Model Type & \bf Acc \\ \hline
Choi, et al.~\shortcite{Choi:2016} & Bi-LSTM-CRF & 67.25 \\
UniTagger-500 & Bi-LSTM & \textbf{79.49} \\
\hline
\end{tabular}
\caption{\label{oov-table} Eojeol-level OOV accuracy}
\end{table}

\subsection{Analysis and Discussion}

Our results show that our model can outperform previous state-of-the-art performance for Eojeol and sentence-level morphological analysis of Korean without linguistic knowledge.

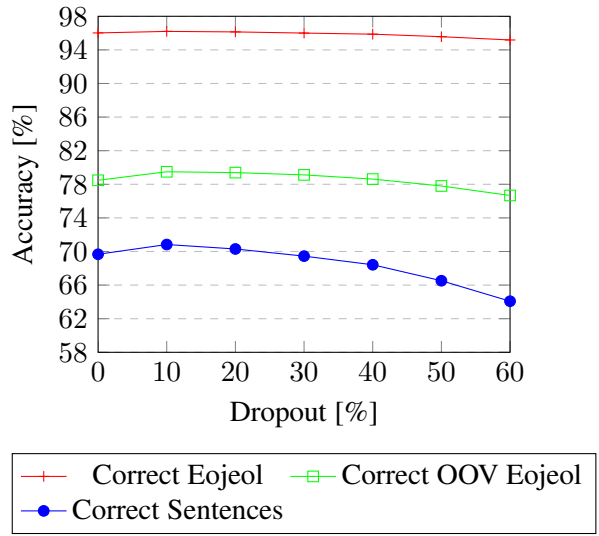
\begin{figure}[!htb]
    \centering
    \begin{tikzpicture}
    \begin{axis}[
        title={},
        xlabel={Dropout [\%]},
        ylabel={Accuracy [\%]},
        xmin=0, xmax=60,
        ymin=58, ymax=98,
        xtick={0,10,20,30,40,50,60},
        ytick={58,62,66,70,74,78,82,86,90,94,98},
        legend style={at={(0.5,-0.3)},
            anchor=north,legend columns=2},
        ymajorgrids=true,
        grid style=dashed,
    ]

    \addplot[
        color=red,
        mark=+,
        ]
        coordinates {
        (0,96.02)(10,96.20)(20,96.14)(30,96.00)(40,95.87)(50,95.57)(60,95.18)
        };
        \addlegendentry{Correct Eojeol}

    \addplot[
        color=green,
        mark=square,
        ]
        coordinates {
        (0,78.48)(10,79.49)(20,79.39)(30,79.12)(40,78.62)(50,77.80)(60,76.67)
        };
        \addlegendentry{Correct OOV Eojeol}

    \addplot[
        color=blue,
        mark=*,
        ]
        coordinates {
        (0,69.67)(10,70.83)(20,70.30)(30,69.45)(40,68.42)(50,66.52)(60,64.08)
        };
        \addlegendentry{Correct Sentences}

    \end{axis}
    \end{tikzpicture}
	\caption{\label{dropout-performance} Impact of dropout on end-to-end tagging performance }
\end{figure}

When the dropout factor is adjusted, all metrics follow a similar trend as seen in Figure~\ref{dropout-performance}. Sentence accuracy is most sensitive to dropout factor adjustment. Best performance is achieved with a dropout rate of 10\%, and increasing the dropout rate further does not increase Eojeol-level OOV accuracy. This is a positive finding, as it indicates the model is not considerably overfitting to the training data beyond approximately the 10\% level.

\begin{figure}[!htb]
    \centering
    \def\svgwidth{\columnwidth}    
    \scalebox{0.70}{\input{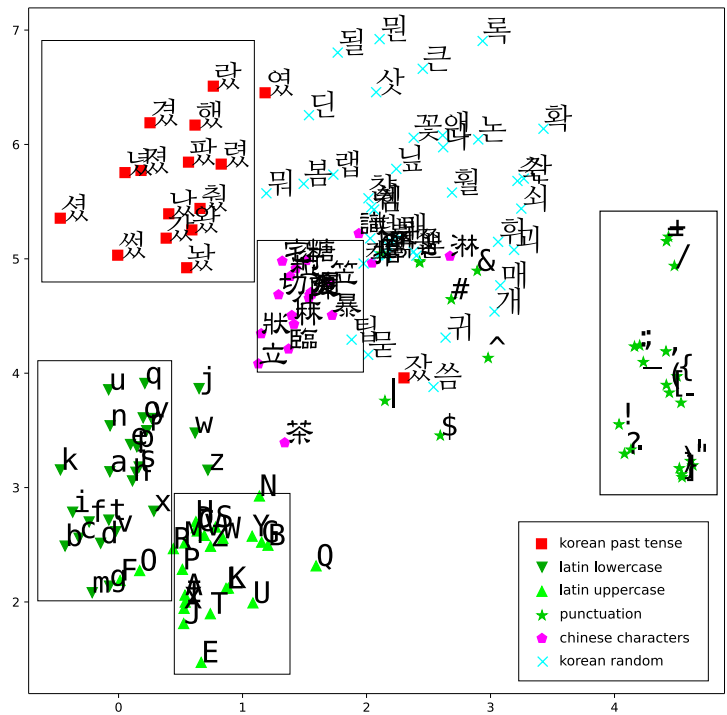}}
	\caption{\label{tsne-morph-embed} Deep embeddings of characters at the morphing stage (t-SNE)}
\end{figure}

In Figure~\ref{tsne-morph-embed}, a 300-dimensional unit embedding layer of the morphing stage is visualized using 2-component t-SNE. The corresponding gold actions are shown in Table~\ref{past-tense-actions}, where all past tense morphemes end with final consonant ``{\begin{CJK}{UTF8}{mj}ㅆ\end{CJK}}'' (ss). The model is able to infer that most of the forms shown in the graph represent the past tense and that they share a similar transformation pattern at the final consonant grapheme level. This shows that our model is able to correlate similar sub-character level morphological transformations even when operating at the character level. Furthermore, it is worth noting that Chinese characters still occur rarely in the Korean language in certain contexts. We can see Chinese characters grouped in a cluster, which shows that the model is able to distinguish one character-rich language (Korean) from another (Chinese). Other characters, such as punctuation, are also grouped by type in largely distinct clusters with occasional overlap.

Joint training of both stages was also attempted, though an initial investigation suggests that performance is not significantly different from training each stage's parameters independently.

The use of using an auxiliary binary break level attribute to represent whitespace was also investigated, but significantly higher accuracy was achieved by using a reserved spacing token instead. Despite the auxiliary break level attribute embedding, both stages of the model have a tendency to learn ambiguous morpheme transformations for adjacent Eojeol. In other words, even though the morpheme transformation and tags are correct, the Eojeol boundaries were incorrectly identified. With the reserved spacing token, this issue was extremely rare.

\section{Conclusion and Future Work}

In this work, we address the commonly held notion that Korean can not be tagged with competitive performance at the character level without prior linguistic knowledge. Our model architecture is not novel compared to previous work. The novelty of our morphological analyzer is its striking simplicity compared to previous approaches for character-rich languages such as Korean. The alignment oracle does not require any cost value for alignment operations such as in the Needleman–Wunsch algorithm~\cite{Needleman:1970}. The Bi-LSTM model is able to learn and utilize alignments that are purely arbitrary and apply them to unseen test data. Even when significantly limiting the number of actions in the training data, we show that by using the most common morphological transformation actions in an agglutinative language, we can exceed the performance of a model that uses linguistic knowledge such as sub-character features. We also show that the widely used CRF layer may in fact be unnecessary for high performance and add unnecessary computational complexity. This exceeded our own expectations and raises the possibility that a single architecture can handle tagging universally with only two simple Bi-LSTM stages. We contribute the necessary source code to replicate the experiments and to attempt alignment and training for any other language, assuming a corpus exists. Nevertheless, there are several points that future work should address.

Out-of-vocabulary morphemes generated by the morphing stage can also result in errors at the tagging stage, as the tagging stage was trained on the assumption of gold morphemes. We would like to experiment with including possible morpheme transformation errors at the tagging stage to determine if tagging performance can be improved.

We attempted joint training but found that end-to-end accuracy was marginally lower. We suspect this is because optimization of each individual stage is hindered by attempting to find optimal parameters for both stages. Future work should attempt joint training of an end-to-end model with the preinitialized parameters from optimizing each stage independently, which has been shown to be ideal in sequential models~\cite{Tang:2016}.

Lastly, although we are unaware of a language more character-rich and more morphologically complex than Korean, we would like to see our model applied to other morphologically complex languages to prove its universality. At the time of writing, we lacked sufficient baseline figures and methodology for generating training data for analyzing other languages at the morpheme level using the Universal Dependencies corpus, and the morphological tags were often conflated with part-of-speech tags. The baselines we found did not specify whether or not morpheme segmentation was taken into account. Without this information, it would be difficult to prove the performance of our model for other languages and we decided to leave training other languages as future work. That being said, our model does not employ any linguistic knowledge specific to Korean, and we therefore have no reason to believe it cannot be trained on any other arbitrary corpus with minor modifications at the preprocessing level.

\section*{Acknowledgements}

This research was supported by the MSIT (Ministry of Science and ICT), South Korea, under the ITRC (Information Technology Research Center) support program ("Research and Development of Human-Inspired Multiple Intelligence") supervised by the IITP (Institute for Information \& Communications Technology Promotion). Additionally, this work was supported by the National Research Foundation of Korea (NRF) grant funded by the South Korean government (MSIP) (No. NRF-2016R1A2B2015912).


\bibliographystyle{acl}
\bibliography{coling2018}

\begin{thebibliography}{}

\bibitem[\protect\citename{Ahn and Seo}2007]{Ahn:2007}
Young-Min Ahn and Young-Hoon Seo.
\newblock 2007.
\newblock Korean part-of-speech tagging using disambiguation rules for
  ambiguous word and statistical information.
\newblock {\em IEEE International Conference on Convergence Information
  Technology}, pages 1598--1601.

\bibitem[\protect\citename{Appleyard \bgroup et al.\egroup
  }2016]{Appleyard:2016}
Jeremy Appleyard, Tomáš Kociský, and Phil Blunsom.
\newblock 2016.
\newblock Optimizing performance of recurrent neural networks on gpus.
\newblock {\em arXiv}, (1604.01946).

\bibitem[\protect\citename{Choi \bgroup et al.\egroup }2016]{Choi:2016}
Jihun Choi, Jonghem Youn, and Sang goo Lee.
\newblock 2016.
\newblock A grapheme-level approach for constructing a korean morphological
  analyzer without linguistic knowledge.
\newblock {\em IEEE International Conference on Big Data}, pages 3872--3879.

\bibitem[\protect\citename{Dong \bgroup et al.\egroup }2016]{Dong:2016}
Chuanhai Dong, Jiajun Zhang, Chengqing Zong, Masanori Hattori, and Hui Di.
\newblock 2016.
\newblock Character-based lstm-crf with radical-level features for chinese
  named entity recognition.
\newblock {\em International Conference on Computer Processing of Oriental
  Languages}, pages 239--250.

\bibitem[\protect\citename{Graves and Schmidhuber}2005]{Graves:2005}
A.~Graves and J.~Schmidhuber.
\newblock 2005.
\newblock Framewise phoneme classification with bidirectional lstm and other
  neural network architectures.
\newblock {\em Neural Networks}, 18(5--6):602--610.

\bibitem[\protect\citename{Heigold \bgroup et al.\egroup
  }2016a]{HeigoldOriginalModel}
Georg Heigold, Guenter Neumann, and Josef van Genabith.
\newblock 2016a.
\newblock Neural morphological tagging from characters for morphologically rich
  languages.
\newblock {\em arXiv}, (1606.06640).

\bibitem[\protect\citename{Heigold \bgroup et al.\egroup
  }2016b]{HeigoldUniversalModel}
Georg Heigold, Guenter Neumann, and Josef van Genabith.
\newblock 2016b.
\newblock Scaling character-based morphological tagging to fourteen languages.
\newblock {\em IEEE International Conference on Big Data}.

\bibitem[\protect\citename{Hochreiter and Schmidhuber}1997]{Hochreiter:1997}
Sepp Hochreiter and Jurgen Schmidhuber.
\newblock 1997.
\newblock Long short-term memory.
\newblock {\em Neural Computation}, 9(8):1735--1780.

\bibitem[\protect\citename{Huang \bgroup et al.\egroup }2015]{Huang:2015}
Zhiheng Huang, Wei Xu, and Kai Yu.
\newblock 2015.
\newblock Bidirectional lstm-crf models for sequence tagging.
\newblock {\em arXiv}, (1508.01991).

\bibitem[\protect\citename{Kang and Kim}1992]{Kang:1992}
Seungshik Kang and Yungtaek Kim.
\newblock 1992.
\newblock A computational analysis model of irregular verbs in korean
  morphological analyzer.
\newblock {\em Journal of Korea Information Science Society}, 19(2):151--164.

\bibitem[\protect\citename{Kingma and Ba}2014]{Kingma:2014}
Diederik Kingma and Jimmy Ba.
\newblock 2014.
\newblock Adam: A method for stochastic optimization.
\newblock {\em arXiv}, (1412.6980).

\bibitem[\protect\citename{Lafferty \bgroup et al.\egroup }2001]{Laffety:2001}
J.~Lafferty, A.~McCallum, and F.~Pereira.
\newblock 2001.
\newblock Conditional random fields: Probabilistic models for segmenting and
  labeling sequence data.
\newblock {\em Proceedings of ICML}.

\bibitem[\protect\citename{Lee and Rim}2005]{Lee1}
Do-Gil Lee and Hae-Chang Rim.
\newblock 2005.
\newblock Probabilistic models for korean morphological analysis.
\newblock {\em Companion to the Proceedings of the International Joint
  Conference on Natural Language Processing}, pages 197--202.

\bibitem[\protect\citename{Lee and Rim}2009]{Lee2}
Do-Gil Lee and Hae-Chang Rim.
\newblock 2009.
\newblock Probabilistic modeling of korean morphology.
\newblock {\em IEEE Transactions on Audio, Speech, and Language Processing},
  17(5):945--955.

\bibitem[\protect\citename{Needleman and Wunsch}1970]{Needleman:1970}
Saul~B. Needleman and Christian~D. Wunsch.
\newblock 1970.
\newblock A general method applicable to the search for similarities in the
  amino acid sequence of two proteins.
\newblock {\em Journal of Molecular Biology}.

\bibitem[\protect\citename{Park \bgroup et al.\egroup }2010]{Park:2010}
Sangwon Park, D~Choi, E~Kim, and K.-S Choi.
\newblock 2010.
\newblock A plug-in component-based korean morphological analyzer.

\bibitem[\protect\citename{Sang and Veenstra}1999]{Sang:1999}
Tjong~Kim Sang and Jorn Veenstra.
\newblock 1999.
\newblock Representing text chunks.
\newblock {\em EACL '99 Proceedings of the ninth conference on European chapter
  of the Association for Computational Linguistics}, pages 173--179.

\bibitem[\protect\citename{Shim and Yang}2004]{Shim:2004}
Gwang-Seob Shim and Jae-Hyung Yang.
\newblock 2004.
\newblock High speed korean morphological analysis based on adjacency condition
  check.
\newblock {\em KIISE: Software and Applications}, 31(1):89--99.

\bibitem[\protect\citename{Tang \bgroup et al.\egroup }2016]{Tang:2016}
Hao Tang, Weiran Wang, Kevin Gimpel, and Karen Livescu.
\newblock 2016.
\newblock End-to-end training approaches for discriminative segmental models.
\newblock {\em IEEE Spoken Language Technology Workshop}.

\end{thebibliography}

\end{document}